# Fusion of Daubechies Wavelet Coefficients for Human Face Recognition

Mrinal Kanti Bhowmik[1], Debotosh Bhattacharjee[2], Mita Nasipuri[2], Dipak Kumar Basu[2*], and Mahantapas Kundu[2]

[1]Department of Computer Science and Engineering, Tripura University
Suryamaninagar- 799130, Tripura, India
Email: mkb_cse@yahoo.co.in

[2]Department of Computer Science and Engineering, Jadavpur University
Kolkata- 700032, India
*AICTE Emeritus Fellow
Email: debotosh@indiatimes.com, {mitanasipuri, dipakkbasu}@gmail.com , mkundu@cse.jdvu.ac.in

*Abstract—* **In this paper fusion of visual and thermal images in wavelet transformed domain has been presented. Here, Daubechies wavelet transform, called as D2, coefficients from visual and corresponding coefficients computed in the same manner from thermal images are combined to get fused coefficients. After decomposition up to fifth level (Level 5) fusion of coefficients is done. Inverse Daubechies wavelet transform of those coefficients gives us fused face images. The main advantage of using wavelet transform is that it is well-suited to manage different image resolution and allows the image decomposition in different kinds of coefficients, while preserving the image information. Fused images thus found are passed through Principal Component Analysis (PCA) for reduction of dimensions and then those reduced fused images are classified using a multi-layer perceptron. For experiments IRIS Thermal/Visual Face Database was used. Experimental results show that the performance of the approach presented here achieves maximum success rate of 100% in many cases.**

## I. Introduction

Many methods have been proposed for face recognition. Fusion of images exploits synergistic integration of images obtained from multiple sensors and by that it can gather data in different forms like appearance and anatomical information of the face, which enriches the system in improving recognition accuracy [9]. As a matter of fact fusion of images has already established its importance in case of image analysis, recognition, and classification. For instance, Aglika Gyaourova et al [10] tried to implemented pixel-based fusion scheme in the wavelet domain, and feature based fusion in the eigenspace domain. Although their fusion approach was not able to fully discount illumination effects present in the visible images but they showed substantial improvements in overall recognition performance. They also indicated that IR-based recognition performance degrades seriously when eyeglasses are present in the probe image but not in the gallery image and vice versa. On the other hand for the improvement of the performance of face recognition when face images are occluded by wearing eye-glasses, Jeong-Seon Park et al [11] first detect the regions occluded by the glasses and generate a natural looking facial image without glasses by recursive error compensation using PCA reconstruction. They proposed a new glasses removal method based on recursive error compensation using PCA reconstruction. George Bebis et al [12] investigated that two different fusion schemes like first one is pixel based and operates in the wavelet domain using Haar transforms, while the second one is feature-based and operates in the eigenspace domain. In both cases, they employ a simple and general framework based on Genetic Algorithms (GAs) to find an optimum fusion strategy. Amit Aran et al [13] demonstrated the spectral band invariant Wave MACH filters which are designed using images of CCD/IR camera fused by Daubechies wavelet transform and implemented in hybrid digital optical correlator architecture to identify multiple targets in a scene .They have fusion of infrared and CCD camera because the performance of CCD camera is better under good illumination conditions where as IR camera gives a better output under poor illumination or in the night conditions also. The authors in [14] proposed data fusion of visual and thermal images using Gabor filtering technique which extracts facial features, are used as a face recognition technique. It has been found that by using the proposed fusion technique Gabor filter can recognize face even with variable expressions and light intensities, but not in extreme condition. Diego A. Socolinsky and Andrea Selinger [15] considered outdoor and indoor imaging conditions for thermal imaging, and one of few to do so even for visible face recognition. It is clear from their experiments that face recognition outdoors with visible imagery is far less accurate than when performed under fairly controlled indoor conditions. For outdoor use, thermal imaging provides us with a considerable performance boost. Thermal recognition performance suffers a moderate decay when performed outside against an indoor enrollment set, probably as a result of environmental changes. Jingu Heo et al [16] describes Comparison results on three fusion-based face recognition techniques like Data fusion of visual and thermal images (Df), Decision fusion with highest matching score (Fh), and Decision fusion with average matching score (Fa) and showed that fusion-based face recognition techniques outperformed individual visual and thermal face recognizers under illumination variations and facial expressions. From them Decision fusion with average matching score consistently demonstrated superior recognition accuracies as per their results. Ioannis Pavlidis and Peter Symosek [17] demonstrated a theoretical and experimental argument that a dual-band (upper and lower band) fusion system in the near infrared can segment human faces much more accurately than traditional visible band disguise face detection systems. Diego A. Socolinsky and Andrea Selinger [18] performed a clear analysis that LWIR imagery of human faces is not only a valid biometric, but almost surely a superior one to comparable visible imagery. Xin Chen, Patrick J. Flynn and KevinW.Bowyer [19] showed that the combination of IR plus visible can outperform either IR or visible alone. They find a combination method that considers the distance values performs better than one that only considers ranks. Christopher K. Eveland et al [20] introduced a methodology for tracking human faces in calibrated thermal infrared imagery of LWIR and MWIR indoor image sequences. H. K. Ekenel and B. Sankur [7] proposed Multiresolution analysis on subspace analysis domain like PCA and ICA. In this work, a technique for human face recognition based on fusion in wavelet transformed domain is proposed and discussed subsequently.

The block diagram of the system is given in Fig 1. All the processing steps used in this paper are shown in the block diagram. In the first step, decomposition of both the thermal and visual images up to level five has been done using wavelet. Then fused image is generated from both the decomposed images. These transformed images separated into two groups namely training set and testing set. The eigenspace is named as fused eigenspaces. Once this projection is done, the next step is to use a classifier to classify them. A multilayer perceptron has been used for this purpose.

### A. Image Decomposition

Wavelet transforms are multi-resolution image decomposition tool that provide a variety of channels representing the image feature by different frequency subbands at multi-scale. It is a famous technique in analyzing signals. When decomposition is performed, the approximation and detail component can be separated [1].

The Daubechies wavelet (db2) decomposed up to five levels has been used here for image fusion. These wavelets are used here because they are real and continuous in nature and have least root-mean-square (RMS) error compared to other wavelets [5] [6].

## II. SYSTEM OVERVIEW

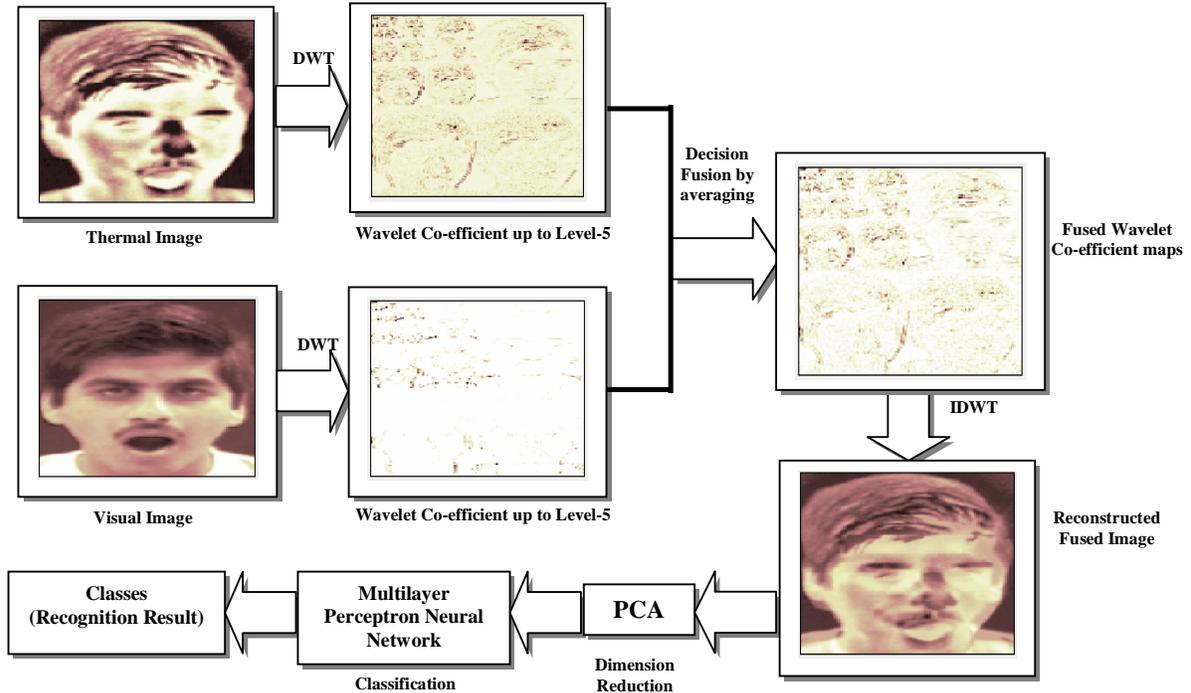

Figure 1: Block diagram of the system presented here.

Daubechies wavelets are a family of orthogonal wavelets defining a discrete wavelet transform and characterized by a maximal number of vanishing moments for some given support. This kind of 2D DWT aims to decompose the image into approximation coefficients (cA) and detailed coefficient cH, cV and cD (horizontal, vertical and diagonal) obtained by wavelet decomposition of the input image (X). The first part of Fig 1 showing after decomposition of two images.

$$[cA, cH, cV, cD] = dwt2 (X, \text{'wname'}) \qquad (1)$$

$$[cA, cH, cV, cD] = dwt2 (X, Lo\_D, Hi\_D) \qquad (2)$$

Equation (1), 'wname' is the name of the wavelet used for decomposition. Equation (2) Lo_D (decomposition low-pass filter) and Hi_D (decomposition high-pass filter) wavelet decomposition filters. This kind of two-dimensional DWT leads to a decomposition of approximation coefficients at level j in four components: the approximation at level j+1, and the details in three orientations (horizontal, vertical, and diagonal). The Fig. 2 describes the algorithmic basic decomposition steps for image where, a block with a down-arrow indicates down-sampling of columns and rows and cA, cH, cV and cD are the coefficient vectors [2] [3] [4].

*B. Image Reconstruction*

The more the decomposition scheme is being repeated, the more the approximation image concentrates in the low frequency energy. To get rid of the illumination effects that may influence the recognition rate, the coefficients in wavelet approximation subband is set to zero. Consequently the reconstruction process is performed using inverse of DWT (IDWT). Finally the reconstruct ted image is used as the input to PCA for recognition.

$$X = idwt2 (cA, cH, cV, cD, \text{'wname'}) \qquad (3)$$

$$X = idwt2 (cA, cH, cV, cD, Lo\_R, Hi\_R) \qquad (4)$$

IDWT uses the wavelet 'wname' to compute the single-level reconstruction of an Image X, based on approximation matrix (cA) and detailed matrices cH, cV and cD (horizontal, vertical and diagonal respectively). By the equation no (4), we can reconstruct the image using filters Lo_R (reconstruct low-pass) and Hi_R (reconstruct high-pass). In the Fig. 3 we have shown the algorithmic basic reconstruction steps for an image.

*C. Principal Component Analysis*

Principal component analysis (PCA) is based on the second-order statistics of the input image, which tries to attain an optimal representation that minimizes the reconstruction error in a least-squares sense. Eigenvectors of the covariance matrix of the face images constitute the eigenfaces. The dimensionality of the face feature space is reduced by selecting only the eigenvectors possessing significantly large eigenvalues. Once the new face space is constructed, when a test image arrives, it is projected onto this face space to yield the feature vector—the representation coefficients in the constructed face space. The classifier decides for the identity of the individual, according to a similarity score between the test image's feature vector and the PCA feature vectors of the individuals in the database [7] [23].

*D. ANN using backpropagation with momentum*

Neural networks, with their remarkable ability to derive meaning from complicated or imprecise data, can be used to extract patterns and detect trends that are too complex to be noticed by either humans or other computer techniques. A trained neural network can be thought of as an "expert" in the category of information it has been given to analyze. The Back propagation learning algorithm is one of the most historical developments in Neural Networks. It has reawakened the scientific and engineering community to the modeling and processing of many quantitative phenomena using neural networks. This learning algorithm is applied to multilayer feed forward networks consisting of processing elements with continuous differentiable activation functions. Such networks associated with the back propagation learning algorithm are also called back propagation networks [8] [21] [22] [23] [24] [25].

### III. EXPERIMENTS RESULTS AND DISCUSSION

This work has been simulated using MATLAB 7 in a machine of the configuration 2.13GHz Intel Xeon Quad Core Processor and 16384.00MB of Physical Memory. We analyze the performance of our algorithm using the IRIS thermal / visual face database.

*A. IRIS Thermal / Visual Face Database*

In this database, all the thermal and visible unregistered face images are taken under variable illuminations, expressions, and poses. The actual size of the images is 320 x 240 pixels (for both visual and thermal). 176-250 images per person, 11 images per rotation (poses for each expression and each illumination). Total 30 classes are present in that database and the size of the database is 1.83 GB [28]. Some fused images of their corresponding thermal and visual images are shown Fig 4.

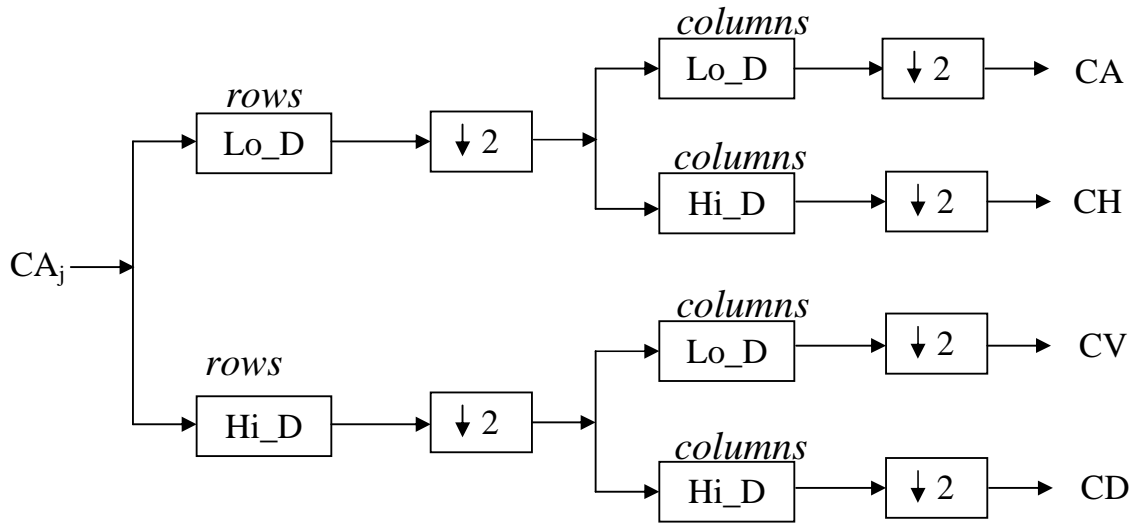

Figure 2: Steps for decomposition of an image

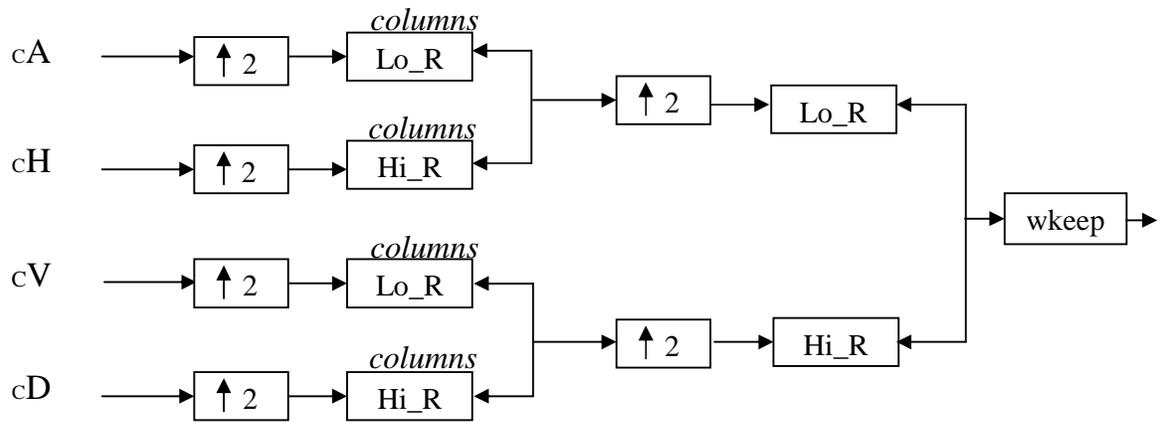

Figure 3. Steps for reconstruction of an image

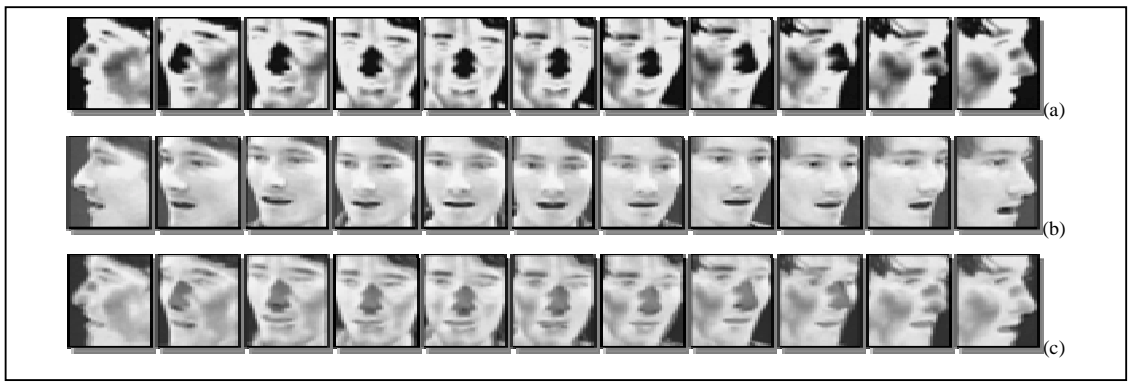

Figure 4. Sample (a) Thermal images (b) Visual images (c) Corresponding Fused images of IRIS database

*B. Training and Testing*

At the time of experiment, we used total 200 visual and 200 thermal images, in which 20 images per class of 10 different classes of IRIS database. Daubechies wavelet transform has been used to generate fused images of both the databases.

The Daubechies wavelet (db2) decomposes the images up to five levels to making fusion image. Here, we consider human face recognition using multilayer perceptron (MLP).

The Daubechies wavelet (db2) decomposes the images up to five levels to making fusion image. Here, we consider human face recognition using multilayer perceptron (MLP). For this research paper, we first train our network using 100 fused images i.e. 10 images per class and those are converted from visual and their corresponding thermal images of IRIS thermal / visual face database. At the time of training, multilayer neural network with back propagation has been used. Momentum allows the network to respond not only to the local gradient, but also to recent trends in the error surface.

After training the network, it was tested with a total of 10 different runs for 10 different classes and all the experiments results of IRIS database are shown in Table I. All these images contained different kind of expressions and 70% of the images were taken in different illumination conditions. The classes with different illuminations with changes in expressions are class–1, class–2, class–3, class–4, class–6, class–7 and class–9, whereas class–5, class–8 and class–10 are with changes in expressions only.

In the Figure 5, all the recognition rates of different classes are presented. From that figure one can observe that the classes, class–3, class–6, class–7 and class-10 are showing highest recognition rate. Out of those four classes, class-3 and class–6 contain the images with changes in illumination as well as expression whereas other two classes contain images with changes is expressions only.

TABLE I.
EXPERIMENTAL RESULTS ON IRIS

| Classes used | No. of Training Images | No. of Testing Images (which are not used during Training) | Recognition Rate |
|---|---|---|---|
| Class - 1 | 10 | 10 | 80% |
| Class – 2 | 10 | 10 | 70% |
| Class – 3 | 10 | 10 | 100% |
| Class – 4 | 10 | 10 | 70% |
| Class – 5 | 10 | 10 | 80% |
| Class – 6 | 10 | 10 | 100% |
| Class – 7 | 10 | 10 | 80% |
| Class – 8 | 10 | 10 | 100% |
| Class – 9 | 10 | 10 | 70% |
| Class – 10 | 10 | 10 | 100% |

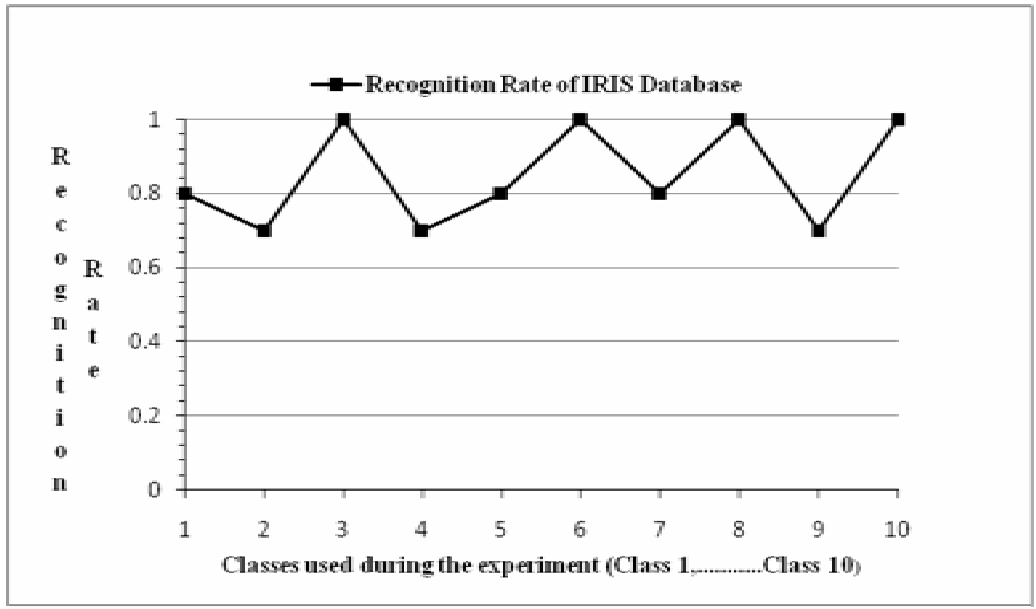

Figure 5. Shows Recognition Rate with False Rejection

TABLE II.
COMPARISON BETWEEN DIFFERENT FUSION TECHNIQUES

| Image Fusion Technique | Recognition Rate |
|---|---|
| Present method | 85% |
| Simple Spatial Fusion[16] | 91.00% |
| Fusion of Thermal and Visual [28] | 90.00% |
| Segmented Infrared Images via Bessel forms [29] | 90.00% |
| Abs max selection in DWT[16] | 90.31% |
| Window base absolute maximum selection [16] | 90.31% |
| Fusion of Visual and LWIR + PCA [17] | 87.87% |

## IV. CONCLUSION

In this a fusion technique for human face recognition using Daubechies wavelet transform on the face images of different illumination with expression has been presented. After completion of fusion, images were projected into an eigenspace. Those projected fused eigenfaces are classified using a Multilayer Perceptron. Eigenspace is constituted by the images belong to the training set of the classifier, which is a multilayer perceptron. The efficiency of the scheme has been demonstrated on IRIS thermal / visual face database which contains images gathered with varying lighting, facial expression, pose and facial details. The system has achieved a maximum recognition rate of 100% in four different cases with an overall recognition rate of 85%.

ACKNOWLEDGMENT

First author is thankful to the project entitled "Development of Techniques for Human Face Based Online Authentication System Phase-I" sponsored by Department of Information Technology under the Ministry of Communications and Information Technology, New Delhi-110003, Government of India Vide No. 12(14)/08-ESD, Dated 27/01/2009 at the Department of Computer Science & Engineering, Tripura University-799130, Tripura (West), India for providing the necessary infrastructural facilities for carrying out this work. The first author would also like to thank Prof. Barin Kumar De of Tripura University, for his kind support to carry out this research work.

REFERENCES

[1] Y. Z. Goh, A. B. J. Teoh, M. K. O. Gog, "Wavelet Based Illumination Invariant Preprocessing in Face Recognition", Proceedings of the 2008 Congress on Image and Signal Processing, Vol. 3, IEEE Computer Society, pp. 421 – 425.
[2] I. Daubechies, Ten lectures on wavelets, CBMS-NSF conference series in applied mathematics. SIAM Ed. 1992.
[3] S. Mallat, "A Theory for Multiresolution Signal Decomposition: The Wavelet Representation", IEEE Transactions on PAMI, 1989, Vol.11, pp. 674 – 693.
[4] Y. Meyer, Wavelets and operators, Cambridge Univ. Press. 1993.
[5] P. Gonzalo, J. M. De La Cruz, "A wavelet-based image fusion tutorial", Pattern Recognition, 2004, Vol. 37, pp. 1855 – 1872.
[6] H. Ma, C. Jia and S. Liu, "Multisource Image Fusion Based on Wavelet Transform", Int. Journal of Information Technology, Vol. 11, No. 7, 2005.
[7] H. K. Ekenel and B. Sankur, "Multiresolution face recognition", Image and Vision Computing, 2005. Vol. 23, pp. 469 – 477.
[8] M. K. Bhowmik, D. Bhattacharjee, M. Nasipuri, D.K. Basu, M. Kundu, "Classification of Polar-Thermal Eigenfaces using Multilayer Perceptron for Human Face Recognition", in Proceedings of the 3rd IEEE Conference on Industrial and Information Systems (ICIIS-2008), IIT Kharagpur, India, December 8-10, 2008, Page no: 118.
[9] I. Pavlidis, P. Symosek, "The imaging issue in an automatic face/disguise detection system", In IEEE Workshop on Computer Vision Beyond the Visible Spectrum: Methods and Applications, pp. 15–24, 2000.
[10] A. Gyaourova, G. Bebis and I. Pavlidis, "Fusion of infrared and visible images for face recognition", Lecture Notes in Computer Science, 2004.Springer, Berlin.
[11] G. Bebis, A. Gyaourova, S. Singh and I. Pavlidis, "Face Recognition by Fusing Thermal Infrared and Visible Imagery," Image and Vision Computing, Vol. 24, Issue 7, July 2006, pp. 727-742.
[12] J. S. Park, Y. H. Oh, S. C. Ahn and S. W. Lee, "Glasses Removal from Facial Image Using Recursive Error Compensation", IEEE Transaction on PAMI, Vol. 27 May 2005.
[13] A. Aran, S. Munshi, V. K. Beri and A. K. Gupta, "Spectral Band Invariant Wavelet-Modified MACH filter" , Optics and Laser in Engineering, Vol. 46, pp. 656-665, 2008.
[14] M. Hanif and U. Ali, "Optimized Visual and Thermal Image Fusion for Efficient Face Recognition", IEEE Conference on Information Fusion, 2006.
[15] D. A. Socolinsky and A. Selinger, "Thermal Face Recognition in an Operational Scenario", IEEE Computer Society Conference on Computer Vision and Pattern Recognition, Vol. 2, pp. 1012-1019, USA, 2004.
[16] J. Heo, S. G. Kong, B. R. Abidi and M. A. Abidi, "Fusion of Visual and Thermal Signatures with Eyeglass Removal for Robust Face Recognition", Conference on Computer Vision and Pattern Recognition Workshop (CVPRW'04) Vol. 8, pp. 122-122, USA, 2004.
[17] I. Pavlidis and P. Symosek, "The Imaging Issue in an Automatic Face/Disguise Detecting System", Proceedings of the IEEE Workshop on Computer Vision Beyond the Visible Spectrum: Methods and Applications (CVBVS 2000).
[18] D. A. Socolinsky and A. Selinger, "A comparative analysis of face recognition performance with visible and thermal infrared imagery," Proceedings of ICPR, Quebec, Canada, August 2002.
[19] X. Chen, P. Flynn and K., Bowyer, "Pca-based face recognition in infrared imagery: Baseline and comparative studies," in IEEE International Workshop on Analysis and Modeling of Faces and Gestures, (Nice, France), 2003.
[20] C. K. Eveland, D. A. Socolinsky and L. B. Wolff, "Tracking Human Faces in Infrared Video", Image and Vision Computing, Vol. 21, 1st July 2003, pp. 579-590.
[21] D. Bhattacharjee, M. K. Bhowmik, M. Nasipuri, D. K. Basu, M. Kundu; "Classification of Fused Face Images using Multilayer Perceptron Neural Network", Procedings of International Conference on Rough sets, Fuzzy sets and

Soft Computing, November 5–7, 2009, organized by Tripura Mathematical Society, School of IT and Computer Science, Tripura University; Indian Society for Fuzzy Math & Information Processing, pp. 289 – 300.
- [22] M. K. Bhowmik, D. Bhattacharjee, M. Nasipuri, D.K. Basu, M. Kundu, "Classification of Polar-Thermal Eigenfaces using Multilayer Perceptron for Human Face Recognition", Proceedings of The 2$^{nd}$ International Conference on Soft computing (ICSC-2008), IET, Alwar, Rajasthan, India, November 8–10, 2008, pp:107-123.
- [23] M. K. Bhowmik, D. Bhattacharjee, M. Nasipuri, D. K. Basu, M. Kundu; " Human Face Recognition using Line Features", proceedings of National Seminar on Recent Advances on Information Technology (RAIT-2009) which will be held on Feb 6-7,2009,Indian School of Mines University, Dhanbad. pp: 385.
- [24] M. K. Bhowmik, "Artificial Neural Network as a Soft Computing Tool – A case study", In Proceedings of National Seminar on Fuzzy Math. & its application, Tripura University, November 25 – 26, 2006, pp: 31 – 46.
- [25] M. K. Bhowmik, D. Bhattacharjee and M. Nasipuri, "Topological Change in Artificial Neural Network for Human Face Recognition", Proceedings of National Seminar on Recent Development in Mathematics and its Application, Tripura University, November 14 – 15, 2008, pp: 43 – 49.
- [26] A. Singh, G. B. Gyaourva and I. Pavlidis, "Infrared and Visible Image Fusion for Face Recognition", Proc. SPIE, vol.5404, pp.585-596, Aug.2004.
- [27] P. Buddharaju, I. T. Pavlidis, P. Tsiamyrtzis, and M. Bazakos "Physiology-Based Face Recognition in the Thermal Infrared Spectrum", IEEE transactions on pattern analysis and machine intelligence, vol.29, no.4, April 2007.
- [28] http://www.cse.ohio-state.edu/otcbvs-bench/Data/02/download.html